\documentclass[letterpaper, 10 pt, conference]{ieeeconf}  
\IEEEoverridecommandlockouts                              
\usepackage{graphics}    
\usepackage{times}       
\usepackage{amsmath}     
\usepackage{amssymb}     
\usepackage{graphicx}
\usepackage{algorithm}
\usepackage[noend]{algpseudocode}
\usepackage{subcaption}
\usepackage{overpic}
\usepackage{siunitx}
\usepackage{url}
\usepackage{hyperref}

 \usepackage[font=small]{caption}

\def\secref#1{Sec.~\ref{#1}}
\def\figref#1{Fig.~\ref{#1}}
\def\tabref#1{Tab.~\ref{#1}}
\def\eqref#1{Eq.~(\ref{#1})}

\newenvironment{hypothesesenum}{
	
	\begin{enumerate}
	}{
	\end{enumerate}
}

\newenvironment{rqenum}{
	
	\begin{enumerate}
	}{
	\end{enumerate}
}

\title{\LARGE \bf 
Auditory Localization and Assessment of Consequential Robot Sounds: \\ A Multi-Method Study in Virtual Reality}

\author{Marlene Wessels \and Jorge de Heuvel \and Leon Müller \and Anna Luisa Maier \and Maren Bennewitz \and Johannes Kraus
  \thanks{M. Wessels, A.L. Maier  \& J. Kraus are with the Johannes Gutenberg-University Mainz, Germany. J. de Heuvel \& M. Bennewitz are with the University of Bonn and the Lamarr Institute for Machine Learning and Artificial Intelligence, Bonn, Germany. L. Müller is with the Chalmers University of Technology, Gothenburg, Sweden.}%
  \thanks{This work has been supported under the grant numbers 16SV9263 and 16KIS1949.}
  \thanks{We thank Thirsa Huisman for helping with the headphone calibration and Mattis Engelsing for data collection. Note that generative AI tools were used  for minor language corrections.
  }%
}

\begin{document}
\maketitle
\thispagestyle{empty} 
\pagestyle{empty}

\begin{abstract} 
    Mobile robots increasingly operate alongside humans but are often out of sight, so that humans need to rely on the sounds of the robots to recognize their presence. For successful human-robot interaction (HRI), it is therefore crucial to understand how humans perceive robots by their consequential sounds, i.e., operating noise. Prior research suggests that the sound of a quadruped Go1 is more detectable than that of a wheeled Turtlebot. 
    This study builds on this and examines the human ability to localize consequential sounds of three robots (quadruped Go1, wheeled Turtlebot 2i, wheeled HSR) in Virtual Reality. 
    In a within-subjects design, we assessed participants' localization performance for the robots with and without an acoustic vehicle alerting system (AVAS) for two velocities (0.3, 0.8 m/s) and two trajectories (head-on, radial). In each trial, participants were presented with the sound of a moving robot for 3~s and were tasked to point at its final position (localization task). Localization errors were measured as the absolute angular difference between the participants' estimated and the actual robot position.
    Results showed that the robot type significantly influenced the localization accuracy and precision, with the sound of the wheeled HSR (especially without AVAS) performing worst under all experimental conditions. Surprisingly, participants rated the HSR sound as more positive, less annoying, and more trustworthy than the Turtlebot and Go1 sound. This reveals a tension between subjective evaluation and objective auditory localization performance. Our findings highlight consequential robot sounds as a critical factor for designing intuitive and effective HRI, with implications for human-centered robot design and social navigation.

\end{abstract}

\vspace{-0.5em}
\section{Introduction}
\vspace{-0.25em}
\label{sec:intro}


    Mobile robots increasingly operate alongside humans in shared spaces such as public spaces and workplaces, where they may not always be visible to humans. In such environments, robot sounds can thus be the only source of information for humans to perceive a robot's presence, and the only source on which to plan actions. Although robots emit mechanical noise when moving, i.e., consequential sound, it is unclear whether and how these sounds - especially different sound profiles originating from different mechanisms of robot movement (e.g., legged vs. wheeled) - influence human auditory perception \cite{allen_consequential_2023, zhang_nonverbal_2023}. This is, however, essential for the design of harmonious, safe, and efficient human-robot interaction (HRI) in shared spaces. 

        \begin{figure}[t]
          \centering
         \includegraphics[width=0.99\linewidth]{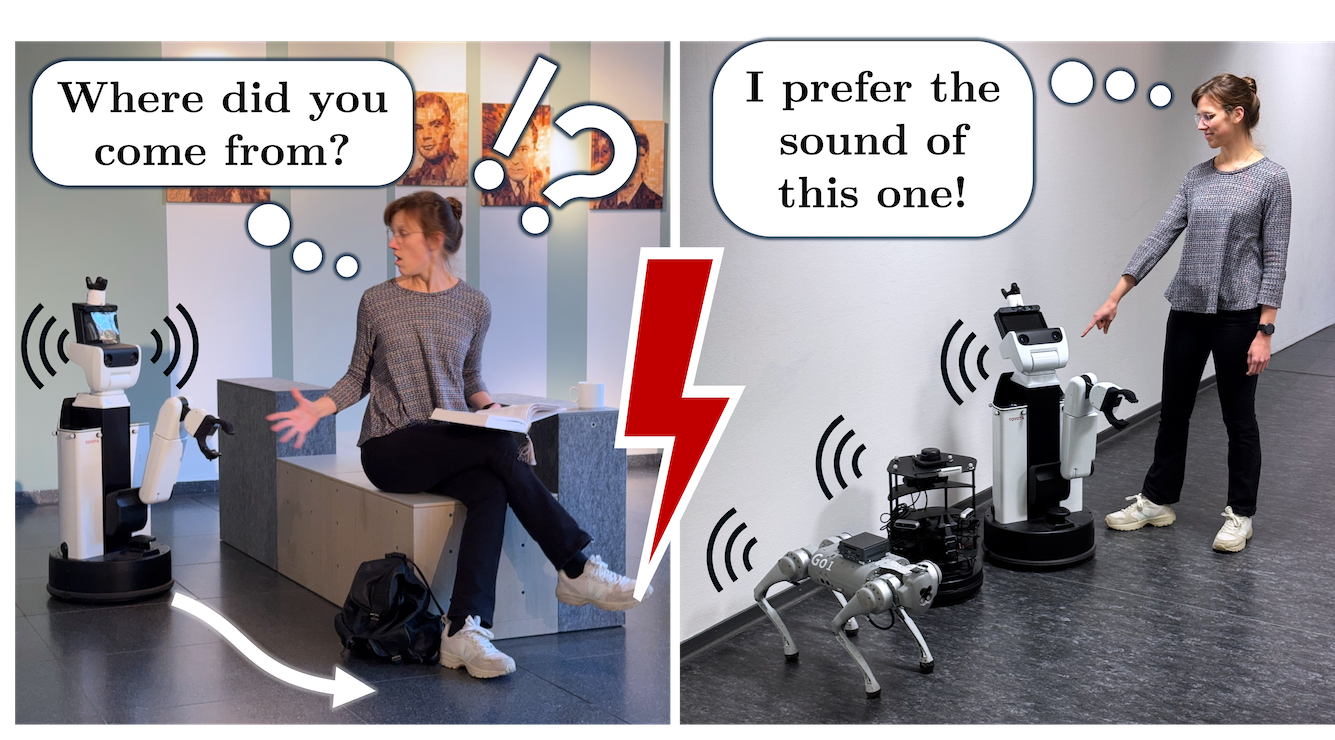}
         \caption{This study examines the impact of consequential sound of three mobile robots on human localization and subjective evaluation. It indicates a tension between objective and subjective measures but consistently highlights consequential robot sounds as key factor for designing a harmonious and effective HRI.}
          \label{fig:motivation}
        \end{figure}

    In shared human-robot environments, a mutual understanding of the current locations of agents and anticipation of future actions are key to smooth coordination and workflow continuity \cite{walch_car-driver-handovers_2017}. 
    Situational awareness (SA) - a psychological variable related to human attention and understanding of situations in which human interact with technology - includes the ability to perceive and understand the environmental context, agent locations (i.e., robot or human), and anticipate future movements \cite{endsley_toward_1995}. 
    While visual cues often dominate SA, the perception of auditory cues such as robot sound becomes crucial when visual information is limited. Effective auditory localization of robot consequential sounds supports humans to detect and track robots, presumably enhancing SA and thus preventing unexpected collisions and unplanned disruptions of actions. Yet, different locomotion types (e.g., wheeled vs. quadruped) generate distinct consequential sounds, which may differently impact human perception and, consequently, the joint action planning and navigation of robots and humans. 
    Despite the potential significance of consequential sounds, there is only limited research on auditory localization for different robotic consequential sounds \cite{zhang_nonverbal_2023, cha_effects_2018}.

    Previous research suggests that quadruped robots, which generate distinct stomping footstep-like sounds, are easier to detect auditorily than wheeled robots, which produce continuous whiny mechanical noise \cite{agrawal_sound_2024}. However, it remains unclear whether this advantage extends to auditory localization of robots. This study therefore aims to address two main research questions:

    \begin{rqenum}
            \item Do different consequential sounds affect the human localization of robots (wheeled vs. quadruped)?
            \item Can the implementation of additional artificial sound improve the localization of robots, similar to acoustic vehicle alerting systems (AVAS) for electric cars~\cite{Mueller2025}?
    \end{rqenum}

    To answer these questions, we conducted a virtual reality (VR) study, which examined participants' accuracy and precision of auditory localization for wheeled and quadruped robots. The robots were either equipped with AVAS or not, and moved at different velocities and trajectories. 

    This study aims at the following contributions:
    \begin{itemize}
        \item Objective evaluation of auditory localization performance for different robot locomotion types (wheeled vs. quadruped) with and without AVAS. 
        \item Measuring the subjective assessment of different robot sounds in terms of annoyance, trust, and valence, offering insights into potential trade-offs between localization effectiveness and user experience.
        \item Deriving implications for human-centered robot sound design and navigation strategies tailored to different robot types to develop robots that better integrate into shared spaces while enhancing safety and efficiency.
    \end{itemize}
    
\vspace{-0.5em}
\section{Related Work}
\vspace{-0.25em}
\label{sec:related}

    Only few studies have investigated the role of consequential robot sounds in human perception and HRI. 
    In an online study by Allen et al. \cite{allen_sound_2025}, two groups of participants ($n$=182) watched videos of five robots either with or without consequential sounds and provided qualitative feedback. Participants largely disliked loud and high-pitched sounds but favored clear and informative sounds (e.g., indicating the robot's state or location) over silence, as these helped to predict the robot’s purpose and trajectory. 
    These results suggest that consequential sounds play an important role in the subjective assessment of robots but leave open the question of how different consequential sounds affect objective auditory perception and situational awareness in HRI.

    Agrawal et al. \cite{agrawal_sound_2024} examined how different types of consequential sounds influence objective auditory  detectability. In a laboratory experiment, 18 participants were tasked to detect a robot sound from background noise in a two-interval forced choice task. The robot sound was generated by a quadruped Go1 and a wheeled Turtlebot 2i and was presented in low and high background noise. The stomping sound of the Go1 was detected at larger distances than the continuous sound of the Turtlebot, even in high background noise. These findings raise the question of whether quadruped robots are not only easier for humans to detect, but possibly also easier to localize than wheeled robots.
    
    One approach to enhance auditory localization of poorly perceptible robots may be AVAS, commonly used in electric mobility (e.g., \cite{Mueller2025}, \cite{wessels_audiovisual_2022}). 
    In an experiment by Cha et al. \cite{cha_effects_2018}, 24 participants were instructed to collaborate with a wheeled Turtlebot 2 in a warehouse scenario and to drop off orders at the robot's location. The robot moved on a straight path towards one of three goal positions. It was hidden behind a curtain, so that participants had to localize the robot by its sound to fulfill the task. The robot was either equipped with a tonal sound, broadband sound, or no AVAS. The localization accuracy generally improved when the robot was equipped with AVAS, compared to the conditions with consequential sounds only. Participants favored broadband sounds over tonal signals as AVAS variant. 
    Thus, the implementation of AVAS may be particularly beneficial for robots that are otherwise difficult for humans to localize, as may be wheeled ones \cite{agrawal_sound_2024}.

    In real-life settings, the directions and motion patterns of robots relative to the position of a human counterpart vary strongly, complicating the predictability of current robot positions. 
    Previous research on representational momentum in spatial hearing indicates that the final position of a moving sound source (i.e., robot) can be perceived as displaced in the direction of their movement \cite{Getzmann.2007}. This suggests that participants might be prone to localization errors for radial robot movements, such as when a robot navigates around them. Since the angular position of a robot on a radial trajectory varies more strongly with a higher than with a lower velocity given a fixed presentation time, we expected the localization performance to be reduced at higher velocities of radial movement.

    Against this background, we hypothesized that...
    \begin{hypothesesenum}
        \item ...the sound of a quadruped robot would be better localized than that of a wheeled robot.
        \item ...an acoustic vehicle alerting system (AVAS) would improve the ability to localize a wheeled robot.
        \item ...robots moving slowly would be better localized than those at a higher speed, at least at a radial trajectory. \vspace{.5mm}
    \end{hypothesesenum}

    As previous studies showed that robot sounds (consequential and AVAS sounds) can impact the subjective evaluation of robots (e.g., \cite{allen_robots_2025, cha_effects_2018, tennent_good_2017, zhang_exploring_2021, babel_findings_2022-1, trovato_sound_2018-1}), we also exploratively examined subjective assessments (annoyance, valence, trust).

\vspace{-0.5em}
\section{Methodology}
\vspace{-0.25em}
\label{sec:meth}
This section outlines the localization task, covering the experimental design, recordings of robot sounds, VR simulation, study procedure, and participant demographics.

\begin{figure*}[!ht]
  \centering
 \includegraphics[width=1\linewidth]{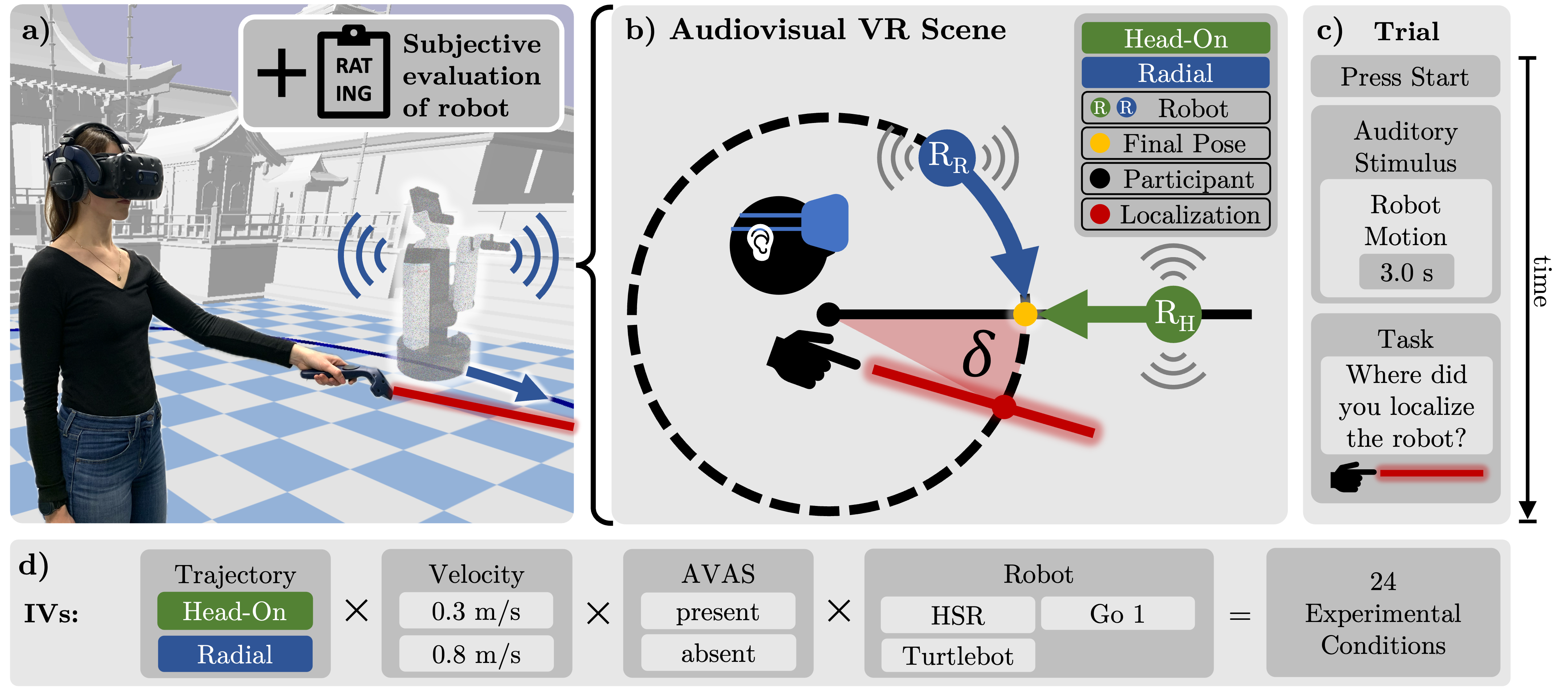}
 \caption{Schematic illustration of experimental study. \textbf{a)} Participant localizes an invisible robot by its sound in VR and subsequently evaluates the different robot sounds. \textbf{b)} Top-down view on audiovisual VR scene. The invisible robot moves towards (head-on trajectory) or around the participant (radial trajectory). For both trajectories, the robot stops at a constant distance of 7~m from the participant and the participant estimates the final pose of the robot. The angular position of the robot was randomized within a span of -90$^\circ$ and +90$^\circ$ (participants’ default orientation was 0$^\circ$). The absolute angular deviation of the estimated from the actual robot pose represents the localization error ($\delta$) in each trial. \textbf{c)} Participants start each trial by pulling the trigger on the controller. The sound of the moving robot is presented for 3.0~s. After the sound has stopped, participants indicate where they last localized the robot and proceed with the next trial. \textbf{d)} Fully-crossed combinations of 4 independent variables (IVs): trajectory, velocity, AVAS and robot (a total of 24 experimental conditions for the localization task, each was repeated 10 times per participant).}
  \label{fig:method}
\end{figure*}

\subsection{Experimental design of the localization task}
    Participants localized the sounds of three robots in a laboratory VR environment.
    The use of an audiovisual VR provided a controlled and immersive setting, which allowed for precise standardization of sensory cues and improved the internal validity of the localization experiment. Specifically, it enabled us to present the exact same robot sound for a specific duration across multiple trials per participant.
    The experiment employed a within-subjects design with four independent variables: robot type (quadruped Go1, wheeled Turtlebot, wheeled HSR), AVAS (with, without AVAS), robot velocity (0.3, 0.8 m/s) and robot trajectory (head-on, radial). A total of 24 conditions (3 robots × 2 AVAS conditions × 2 velocities × 2 trajectories) was tested. Each condition was repeated 10 times per participant (240 trials total). The two trajectories were varied in blocks and their order was counterbalanced across participants. Within each block, we presented the respective combinations of robot type, AVAS and velocity in randomized order.
    
    In the head-on block, the robot approached the participant directly and stopped at 7~m. In the radial block, it circled the participant at a constant distance of 7~m.
    The final robot distance was 7~m in both blocks, with its angular position randomized between  -90$^\circ$ and +90$^\circ$ relative to the participant’s forward orientation  (0$^\circ$). Note that for radial trials, the robot moved to the left when the final angle was -90$^\circ$ and to the right when the final angle was +90$^\circ$. Otherwise, the movement direction was randomized.

\subsection{Recordings and simulation of robot sounds} \label{sec:sounds}

    Participants were presented with sounds of moving robot in VR. The stimulus sounds were based on acoustic recordings of three real robots: Go1, Turtlebot 2i and HSR. These recordings were conducted in an office hallway by mounting two calibrated GRAS 64AE 1/2"  microphones, a camera, and a HeadAcoustics SQobold mobile data acquisition system on each robot. To reduce the influence of the acoustic environment, the microphones were positioned closely to the parts of the robots that were aurally identified as dominant sound sources. Then, each robot was recorded while moving at multiple constant and accelerating trajectories in the hallway. To obtain an estimate of the robot velocities, the camera mounted on each robot was pointed toward a measuring tape on the floor and captured synchronously with the microphone signals. These video recordings were then manually analyzed by marking the robot position in 1~m increments to derive velocity estimates. Additionally, binaural reference recordings of the robots moving toward a listener position were conducted with a calibrated HeadAcoustics BHS II headset, using the same camera-based approach to estimate the distance between robot and listener.

    \begin{figure}[t]
    \centering
     \includegraphics[width=1\linewidth]{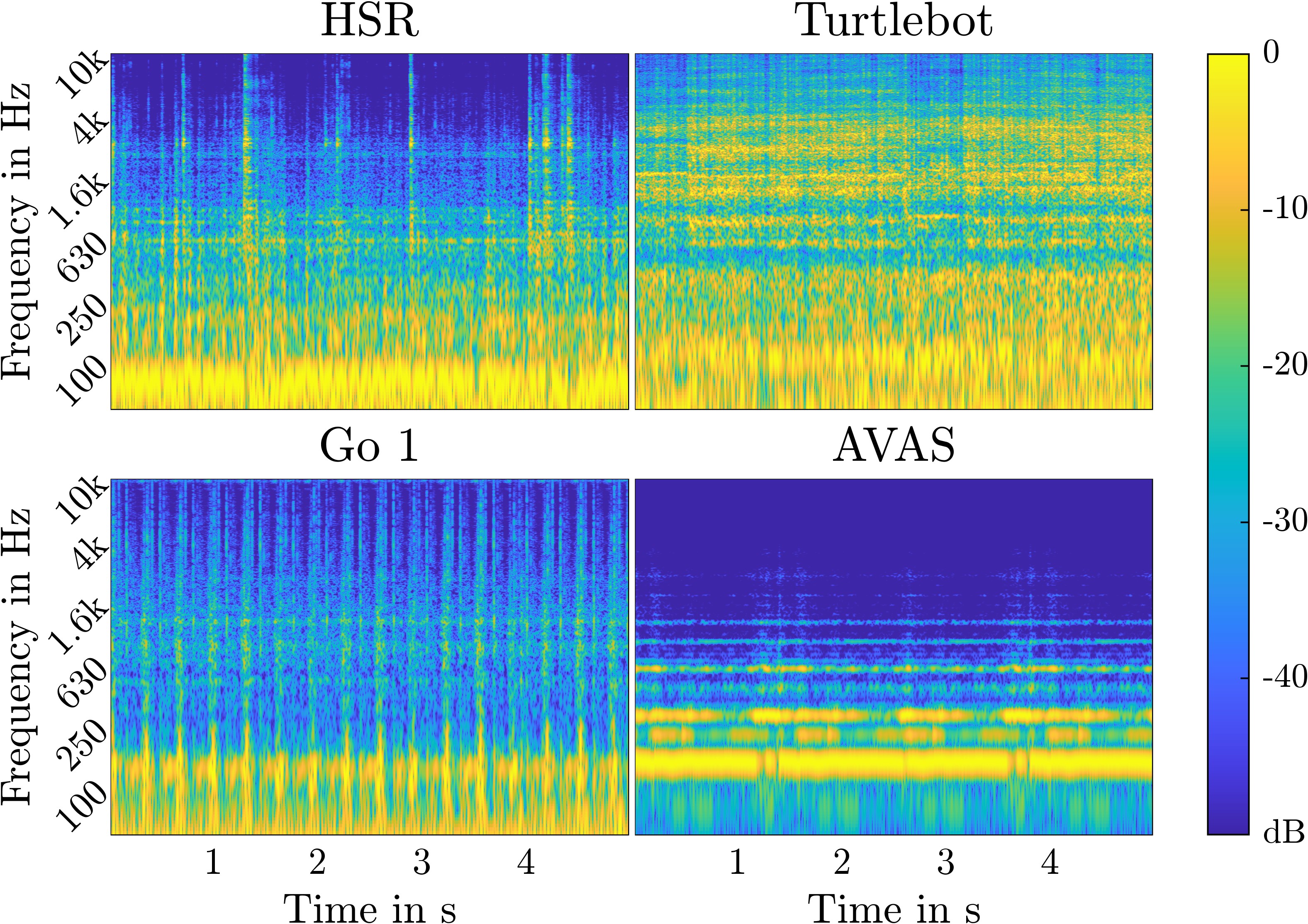}
     \caption{Normalized sound pressure spectrograms of the wheeled HSR (top left), wheeled Turtlebot 2i (top right), quadruped Go1 (bottom left), and isolated AVAS signal (bottom right). In the study, the AVAS signal was combined with the three robot sounds.}

    \label{fig:spectrogtams}
    \end{figure}  
    
    All recordings were processed by manually selecting the microphone that yielded the subjectively most plausible sound, removing interfering noises, partitioning the remaining data into overlapping blocks of 5~s, and calculating the mean velocity and velocity variance for each block. Although all evaluated robots had different operating speed ranges, analyzing the distribution of these velocity values showed that sufficiently many usable 5~s recording blocks at a constant speed of 0.35~m/s were obtained for all three robots. Therefore, we selected these recordings and exported 10 different 5~s blocks for each robot to allow for variation in the stimulus set. Note that the velocity of the simulated sound source  movement (i.e., robot) in the virtual environment was set independently of the robots' velocity during the recordings.
    
    \figref{fig:spectrogtams} shows normalized spectrograms for the stimulus sounds of the three robots (sounds available online\footnote{Sounds can be retrieved from \href{https://doi.org/10.5281/zenodo.15064329   }{https://doi.org/10.5281/zenodo.15064329}}).
    
   The spectrograms in \figref{fig:spectrogtams} reveal that the three robots radiate fundamentally different types of sounds: The HSR emitted a soft low-frequency humming noise with some occasional clicks from the steering motors, reaching an equivalent sound pressure level (SPL) of 57~dBA measured at approx. 10~cm distance to the robot's base. The Turtlebot emitted a broadband noise combined with a tonal fluctuating whirring sound caused by the gear transmission, reaching a SPL of 72~dBA measured at approximately 10~cm distance to the robot's base. The Go1 radiated an impulsive impact sound for every step combined with a high-frequency electric motor whirring, reaching an equivalent SPL of 74~dBA measured at approximately 15~cm distance. Apart from the isolated consequential robot sounds, we presented the robot sounds with an added AVAS sound. This AVAS sound was obtained from the commercial Adobe Stock database and comprised several amplitude-modulated tonal components as shown in \figref{fig:spectrogtams}, resembling a hovering sound. It was added with a level of 70~dBA to all robot recordings. Additionally, outdoor background noise was recorded by placing an EigenMike32 spherical microphone array on a quiet parking lot. The noise recording included nature sounds and distant traffic noise but no salient auditory events\textsuperscript{1}. The noise recording was included in the VR simulation as a 1\textsuperscript{st}-order Ambisonics rendering and played back with a SPL of approx. 40~dBA.

    The sound propagation from the robot to the participant was modeled using the software TASCAR \cite{grimm_toolbox_2019}. TASCAR adjusted the sound simulation depending on the dynamic positions of the robot and the participant's head, so that it accounted for the distance-dependent changes in sound level and propagation time. Similar to a real scene, the acoustic simulation included intensity and spectral changes, and interaural time and level differences as auditory cues for the robot's motion. Participants perceived the dynamic spatial sound field via headphones (Beyerdynamic DT 770 Pro, 80 Ohm, calibrated to 80~dBA). 
    
    Studies such as \cite{Yost.2016} suggest that sound localization accuracy does not depend on overall sound pressure level as long as the sound is detectable. Therefore, we decided to reduce level differences within the experiment by normalizing all robot sounds without AVAS to an equivalent sound pressure level of 80~dBA after the binaural rendering at a 7~m distance. By doing so, we ensured that a potential effect of robot type is not driven by systematic level differences between robots.
    

\subsection{Simulation of visual environment}
    Participants experienced a visual virtual environment via an HTC Vive Pro 2 head-mounted display (HMD). 
    The virtual scene was developed with the VR version of PyBullet \cite{coumans_pybullet_2016}. 
    To enable real-time 3D~auditory simulation via the headphones worn by the participant, the HMD’s pose data was continuously streamed to TASCAR via the Open Sound Control protocol.
    As a result, the robot sounds were perceived as spatially anchored within the virtual scene.

    The visual environment depicted a spacious open area with a checkerboard floor and a surrounding 3D model of an inner courtyard to aid participants’ spatial orientation (\figref{fig:method}). 
    An arrow on the floor marked the intended participant position in the center (0, 0) in alignment with the desired participant default orientation at the beginning of each trial.
    Additionally, we depicted a blue circle with a radius of 7~m on the checkerboard floor. 
    It corresponded to the distance of the invisible robot for radial trajectories and to the final distance of the invisible robot for head-on trajectories, and was instructed as such to the participants to allow for a rough spatial representation of the scene.
    Once a participant started a trial, this circle changed its color to green.
    The handheld controller projected a straight beam for the participant to use it as a pointer in the localization task.

\subsection{Auditory localization task}
\label{sec:LocalTask}
    Participants started each trial by pulling the controller's trigger. The sound of a moving robot was presented for~3~s. The simulated robot was audible but never visible. Participants were instructed to localize the robot based on its sound and to use a hand-held controller to point in the direction of the robot's final position after the sound stopped. To confirm their estimate of the robot's position, participants pulled the trigger on the controller again. The absolute angular deviation (in degrees) between estimated and actual robot position was recorded as the localization error for each trial. After each response, participants were instructed to face again in the default direction indicated by the simulated arrow (pointing at angular position 0$^\circ$).


\subsection{Procedure}
    Before the study, participants were provided with written and verbal information about the experiment and gave written consent for voluntary participation, data processing, and publication. Afterwards, they completed demographic questions, and the experimenter measured their interpupillary distance and optimized the alignment of the HMD displays. The experimenter instructed the participants for the first block of the localization experiment (half of the participants started with the head-on block, the rest with the radial block).

    Prior to the experiment and after every 30 localization trials, participants rated their motion sickness levels using the Fast Motion Sickness Scale \cite{keshavarz_validating_2011} to monitor potential symptoms. This scale ranged from “not sick at all” (0) to “frank sickness” (20). Participants also completed 10 training trials (5 for each trajectory) to familiarize themselves with the VR environment and the localization task.
    Then, they proceeded with the 240 experimental trials (see \secref{sec:LocalTask}). The experimenter offered participants a break after every 60 trials and explicitly asked them to rest after completing the first block. For the second block, the experimenter explained that the trajectory of the robot would change (to radial or head-on trajectory, respectively). 
    
    Subsequently, participants evaluated six sounds (consequential sounds of the three robots with and without AVAS) in three separate rating blocks. The first rating block assessed the perceived annoyance of the sounds (11-point ICBEN scale \cite{fields_standardized_2001}, 0 = not at all annoyed by the sound, 10~=~extremely annoyed by the sound), the second block assessed valence (9-point SAM scale \cite{bradley_measuring_1994}, 1 = negative, 5 = neutral, 9~=~positive), and the third block assessed trust (7-point Likert-scale, "I trust the robot." 1 = I do not agree at all, 7 = I agree entirely) adapted from \cite{koverola_general_2022}.

    The study concluded with open questions regarding participants' personal experiences with robots in general and in course of the experiment. The entire session lasted approximately 1.5 hours.
    
\subsection{Participants}
    As preregistered (AsPredicted \#198036), we recruited a sample of $n$=24 participants. They were on average \textit{M\textsubscript{age}}=26.42 years old (\textit{SD\textsubscript{age}}=9.04 years; 13 women, 11 men; 21 right-handed, 3 left-handed). Individuals with a history of seizure disorder were excluded from participation, as VR headsets can potentially trigger seizures. All participants reported (corrected-to-)normal vision and hearing. On average, they indicated limited experience with robots (\textit{M\textsubscript{exp}}=1.75, \textit{SD\textsubscript{exp}}=1.45) on a 7-point rating scale (0 = no experience, 6 = extensive experience). 
    
    The study was conducted in accordance with the ethical principles of the Declaration of Helsinki and was approved by the Ethics Committee of the Institute of Psychology at Johannes Gutenberg-University Mainz (approval number: 2024-JGU-psychEK-S062). Participation was voluntary and was compensated with course credit.

\vspace{-0.5em}
\section{Results and Discussion}
\vspace{-0.25em}
\label{sec:res}
    Prior to the analyses, we excluded extreme data points of the localization task according to a Tukey criterion per combination of participant and the 24~experimental conditions~\cite{tukey_exploratory_1977}. In sum, 94 trials (1.63\%) were detected as extreme data points with three interquartile ranges below the first or above the third quartile. We subsequently aggregated the localization data per combination of participant and experimental condition. In this, we derived means (localization \textit{accuracy}) and standard deviations of the absolute localization errors (localization \textit{precision}) as performance measures. Accuracy reflects how close estimations were to the actual robot position. Precision captures the consistency of responses per participant across trials and thus indicates participants' estimation certainty \cite{fechner_elemente_1860}. For both measures, lower localization error values ($M$ or $SD$) indicate higher accuracy or precision, respectively.
    
    We analyzed the two localization performance measures separately for the trajectory blocks (head-on in \secref{sec:res_headon}, radial in \secref{sec:res_radial}). For each of the four combinations of localization measure and trajectory, we conducted a repeated-measures (rm)ANOVA with the factors \textit{velocity}, \textit{AVAS} and \textit{robot}, using a univariate approach and Huynh-Feldt sphericity correction for the degrees of freedom. We followed up significant main effects of the robot, using pairwise $t$-tests with Bonferroni-correction. In accordance with the specific direction of the effect in H1, we used one-sided $t$-tests for the comparisons of wheeled and quadruped robots and two-sided tests for the comparisons between the two wheeled robots, as for the latter comparison no specific direction was hypothesized. The alpha-level was set to 5\%.
    
    Note that we preregistered an analysis that included trajectory as a factor. However, in this article, we present separate analyses for the two trajectory blocks. This decision aligns with the experimental design and procedure, as the two trajectories were presented in distinct blocks with explicit instructions for each of them.

\subsection{Localization performance for head-on trajectories}

\label{sec:res_headon}
    Participants localized the robots approaching on head-on trajectories with relatively high accuracy and showed, on average, an absolute localization error of 10.52$^\circ$ ($SD$=7.64$^\circ$). As shown in \figref{fig:spectrogtams}a), the mean absolute localization error was descriptively largest for the wheeled HSR ($M$=13.15$^\circ$, $SD$=11.24$^\circ$), followed by the quadruped Go1 ($M$=9.76$^\circ$, $SD$=6.73$^\circ$), and the wheeled Turtlebot ($M$=8.64$^\circ$, $SD$=5.80$^\circ$). 
    
    The rmANOVA confirmed a significant effect of robot type on localization accuracy (results in \tabref{tab:results_headon}). The follow-up tests revealed that participants localized the HSR significantly less accurately than the Turtlebot ($t$(23)=3.36, $p_{\textit{Bonf}}$=.008, $dz$=0.69) and the Go1 ($t$(23)=2.58, $p_{\textit{Bonf}}$=.025, $dz$=0.53), whereas the comparison between Turtlebot and Go1 ($t$(23)=2.07, $p_{\textit{Bonf}}$=.300, $dz$=0.42) was not significant. Neither velocity, AVAS, nor any interaction had a significant impact on localization accuracy.
    
    \begin{figure*}[t]
      \centering
      \begin{subfigure}[t]{0.49\linewidth}
        \centering
        \includegraphics[width=\linewidth]{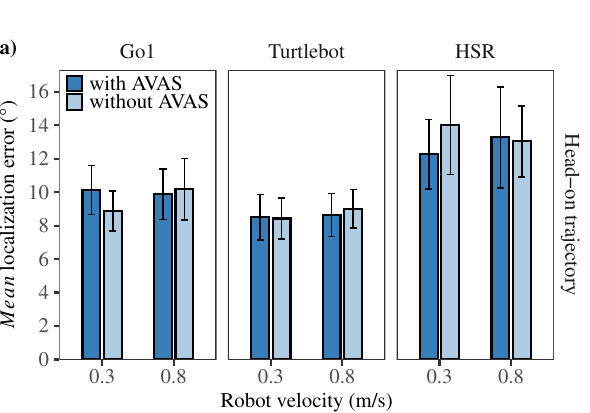}
        \label{fig:resError_headon}
      \end{subfigure}
      \hfill
      \begin{subfigure}[t]{0.49\linewidth}
        \centering
        \includegraphics[width=\linewidth]{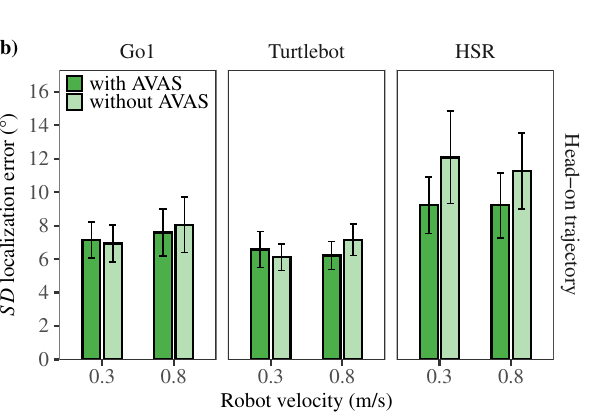}
        \label{fig:res_SD_Error_headon}
      \end{subfigure}
        \vspace{-1.5em}
      \caption{Localization performance for \textit{head-on trajectories} as a function of velocity, AVAS, and robot. \textbf{a)} Localization \textit{accuracy} as measured by the mean absolute localization error (blue). \textbf{b)} Localization \textit{precision} as measured by the intra-individual standard deviation (SD) of the absolute localization error (green). Error bars indicate ±1 $SE$ of the 24 individual values. The descriptive pattern consistently indicates a poorer performance for the HSR (particularly without AVAS) compared to the Turtlebot and Go1.}
      \label{fig:combined_headon}
    \end{figure*}

    Regarding the localization precision, participants varied their estimates on average by 8.13$^\circ$ ($SD$=5.90$^\circ$) in head-on trials. The localization precision differed significantly between robots (\figref{fig:combined_headon}b) and \tabref{tab:results_SD_headon}). Participants localized the wheeled HSR significantly less precisely ($M$=10.44$^\circ$, $SD$=9.26$^\circ$) than the wheeled Turtlebot ($M$=6.51$^\circ$, $SD$=4.06$^\circ$, $t$(23)=2.97, $p_{\textit{Bonf}}$=.021, $dz$=0.61) and the quadruped Go1 ($M$=7.43$^\circ$, $SD$=5.38$^\circ$, $t$(23)=2.69, $p_{\textit{Bonf}}$=.020, $dz$=0.55). The difference between Go1 and Turtlebot was not significant ($t$(23)=1.54, $p_{\textit{Bonf}}$=.828, $dz$=0.31). 
    
    The descriptive pattern in \figref{fig:combined_headon}b) reveals that participants localized the HSR more precisely with AVAS ($M$=9.21$^\circ$, $SD$=8.70$^\circ$) than without ($M$=11.67$^\circ$, $SD$=10.87$^\circ$). This impact of AVAS was, however, not consistently observed across robots (no significant effect of AVAS, contrary to H2). Moreover, neither the interaction between robot type and AVAS nor any other effect was significant (\tabref{tab:results_SD_headon}). \\

    In sum, these findings suggest that the robot type significantly influences localization accuracy and precision in \textit{head-on trajectories}, with the wheeled HSR consistently leading to poorer performance (particularly without AVAS) compared to the Turtlebot and Go1 robots. Since the HSR, rather than wheeled robots in general, was localized less accurately than the quadruped Go1, H1 was only partially confirmed. AVAS and velocity played a minor role for robots moving straight toward participants (H2, H3 not statistically supported). 

    \begin{table}[t]
    \centering
    \begin{tabular}{rcccccc}
    \hline
    \textit{\textbf{Predictor}} & \textit{\textbf{df}} & \(\mathit{\mathbf{\tilde{\epsilon}}}\) & \textit{\textbf{F}} & \textit{\textbf{p}} & \(\mathit{\mathbf{\eta^2_p}}\) \\
    \hline
    Velocity                   & 1,23 &       &  0.53 & .476 & .02 \\
    AVAS                       & 1,23 &       &  0.10 & .755 & .00 \\
    \textbf{Robot} & \textbf{2,46} & \textbf{0.65}  & \textbf{8.66} & \textbf{.004} & \textbf{.27} \\
    Velocity \( \times \) AVAS        & 1,23 &       &  0.00 & \>.99 & .00 \\
    Velocity \( \times \) Robot       & 2,46 & 0.70  &  0.08 & .854 & .00 \\
    AVAS \( \times \) Robot           & 2,46 & 0.87  &  0.39 & .652 & .02 \\
    Velocity \( \times \) AVAS \( \times \) Robot & 2,46 & 0.61  &  0.73 & .428 & .03 \\
    \hline
    \end{tabular}
    \caption{Results of rmANOVA on head-on localization \textit{accuracy}. Significant effect highlighted in bold font.}
    \label{tab:results_headon}
    \end{table}

\subsection{Localization performance for radial trajectories}

\label{sec:res_radial}
    In radial trials, participants localized the robots with an average accuracy of 10.33$^\circ$ ($SD$=7.72$^\circ$), which is comparable to head-on trials. The mean absolute localization error (\figref{fig:spectrogtams}a)) was descriptively larger for the HSR ($M$=12.40$^\circ$, $SD$=9.58$^\circ$) than for the Go1 ($M$=9.39$^\circ$, $SD$=7.05$^\circ$) and the Turtlebot ($M$=9.22$^\circ$, $SD$=6.84$^\circ$).
    
    The rmANOVA confirmed a significant and large effect of robot type on the mean localization error (results in \tabref{tab:results_radial}). The follow-up tests revealed that participants localized the HSR significantly less accurately than both the Go1 ($t$(23)=3.98, $p_{\textit{Bonf}}$=.001, $dz$=0.81) and the Turtlebot ($t$(23)=4.29, $p_{\textit{Bonf}}$=.001, $dz$=0.88). There was no significant difference in localization accuracy between the Turtlebot and Go1 ($t$(23)=0.68, $p_{\textit{Bonf}}>$.999, $dz$=0.14). 
    
    Furthermore, participants localized robots on a radial trajectory with a significantly higher accuracy when they moved at a lower ($M$=9.87$^\circ$, $SD$=7.87$^\circ$) than at a higher velocity ($M$=10.80$^\circ$, $SD$=7.71$^\circ$). Note that with a lower velocity, a robot covered a shorter distance within the 3~s presentation time than with a higher velocity. Thus, the angular position of the slower robot changed comparably less throughout a trial. This implies that the chance for a higher localization performance was higher for a slower robot. This might explain why velocity played a substantial role in radial but not in head-on trials, as in the latter case the angular position of the robot never changed during the presentation time, independent of the robot's velocity.

    \begin{figure*}[t]
      \centering
      \begin{subfigure}[t]{0.49\linewidth}
        \centering
        \includegraphics[width=\linewidth]{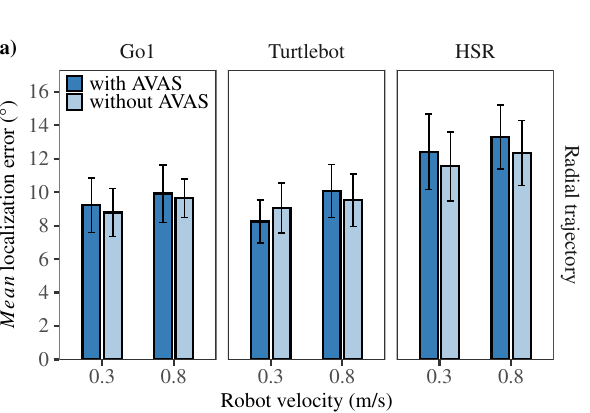}
        \label{fig:resError_radial}
      \end{subfigure}
      \hfill
      \begin{subfigure}[t]{0.49\linewidth}
        \centering
        \includegraphics[width=\linewidth]{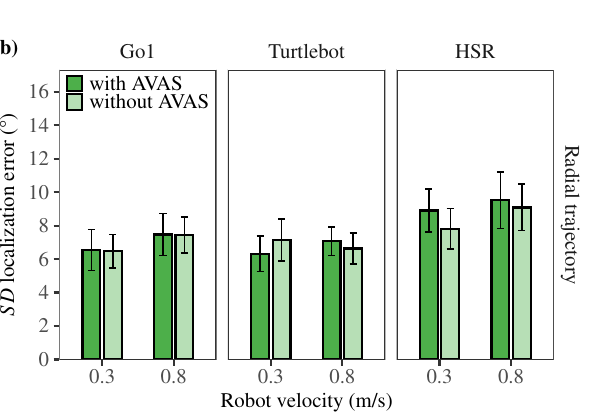}
        \label{fig:res_SD_Error_radial}
      \end{subfigure}
      \vspace{-1.5em}
      \caption{Localization performance for \textit{radial trajectories} as a function of velocity, AVAS, and robot. \textbf{a)} Localization accuracy as measured by the mean absolute localization error (blue). \textbf{b)} Localization precision as measured by the intra-individual standard deviation of the absolute localization error (green). Error bars indicate ±1 $SE$ of the 24 individual values.  The descriptive pattern consistently indicates a poorer performance for the HSR than for the Turtlebot and Go1, as well as for the higher than for the lower velocity.}
      \label{fig:combined_radial}
    \end{figure*}

    \begin{table}[t]
    \centering
    \begin{tabular}{rccccc}
    \hline
    \textit{\textbf{Predictor}} & \textit{\textbf{df}} & \(\mathit{\mathbf{\tilde{\epsilon}}}\) & \textit{\textbf{F}} & \textit{\textbf{p}} & \(\mathit{\mathbf{\eta^2_p}}\) \\
    \hline
    Velocity                   & 1,23 &       & 0.26  & .616 & .01 \\
    AVAS                       & 1,23 &       & 2.33  & .141 & .09 \\
    \textbf{Robot}             & \textbf{2,46} & \textbf{0.68}  & \textbf{7.55} & \textbf{.006} & \textbf{.25} \\
    Velocity \( \times \) AVAS & 1,23 &       & 0.15  & .700 & .01 \\
    Velocity \( \times \) Robot & 2,46 & 0.72  & 0.47  & .567 & .02 \\
    AVAS \( \times \) Robot    & 2,46 & 1.05  & 1.81  & .174 & .07 \\
    Velocity \( \times \) AVAS \( \times \) Robot & 2,46 & 0.62  & 0.33  & .618 & .01 \\
    \hline
    \end{tabular}
    \caption{Results of rmANOVA on head-on localization \textit{precision}. Significant effect highlighted in bold font.}
    \label{tab:results_SD_headon}
    \end{table}

    The localization precision in radial trials was with an average intra-individual variation of 7.53$^\circ$ ($SD$=5.24$^\circ$) on a similar level as in head-on trials. The localization precision differed significantly between robots (results in \tabref{tab:results_SD_radial}, \figref{fig:spectrogtams}b)). 
    The largest variation and thus the lowest precision resulted for the HSR ($M$=8.83$^\circ$, $SD$=6.33$^\circ$), which was significantly lower than both for the Go1 ($M$=6.98$^\circ$, $SD$=5.22$^\circ$, $t$(23)=2.68, $p_{\textit{Bonf}}$=.020, $dz$=0.55) and the Turtlebot ($M$=6.79$^\circ$, $SD$=4.73$^\circ$, $t$(23)=3.70, $p_{\textit{Bonf}}$=.004, $dz$=0.76). The localization precision did not vary significantly between Go1 and Turtlebot ($t$(23)=0.50, $p_{\textit{Bonf}}>$=.999 $dz$=0.10).
    
    The variation in localization estimates was significantly lower and thus participants' localization was more precise for slower ($M$=7.19$^\circ$, $SD$=5.29$^\circ$) than for faster robots ($M$=7.87$^\circ$, $SD$=5.29$^\circ$), likely due to greater angular variation of the robot in radial trajectories at higher speeds, which introduced more positional uncertainty.
 \\

    Taken together, these findings indicate that the robot type plays a crucial role in the localization accuracy and precision for \textit{radial trajectories}. As in head-on trials, participants showed a substantially poorer performance when localizing the wheeled HSR compared to the wheeled Turtlebot and quadruped Go1 (partial support of H1). In other words, participants were more accurate and more certain when they localized the Go1 and Turtlebot, independent of the trajectory, AVAS, or velocity of the robot. For radial trajectories, our data confirmed that slower-moving robots were better localized than those at a higher speed (supporting H3), likely due to reduced angular displacement during trials. 
    
    Considering the time-frequency structure of the different robot sounds (c.f. \figref{fig:spectrogtams}), this outcome is in line with psychoacoustic literature suggesting that broad spectrum sounds, such as the Turtlebot noise, and impulsive sounds, such as the Go1 noise, are easier to localize than steady state low-frequency noise like the HSR sound \cite{Carlini.2024}. Even though the HSR localization results show a trend to slightly improve with added AVAS, these differences were not found to be significant (not supporting H2).
    In this study, we normalized all robot sounds to the same playback level (see \secref{sec:sounds}). However, the potential effect of an AVAS could become significant if all sounds are played back at their original level. While previous studies have shown that added AVAS can significantly increase the detectability of e.g. quiet electric vehicles that would otherwise not be noticed at a safe distance \cite{Emerson.2013}, recent research suggests that different AVAS signals have different effects on localization \cite{Mueller2025}. The multi-tone hover sound may not have been appropriate for the three robots used in this study. Further research should investigate the fit between robot types and different AVAS variants.

\begin{table}[t]
    \centering
    \begin{tabular}{rccccc}
    \hline
    \textit{\textbf{Predictor}} & \textit{\textbf{df}} & \(\mathit{\mathbf{\tilde{\epsilon}}}\) & \textit{\textbf{F}} & \textit{\textbf{p}} & \(\mathit{\mathbf{\eta^2_p}}\) \\
    \hline
    \textbf{Velocity}  & \textbf{1,23} &       &  \textbf{4.88} & \textbf{.037} & \textbf{.18} \\
    AVAS                       & 1,23 &       &  2.52 & .126 & .10 \\
    \textbf{Robot} & \textbf{2,46} & \textbf{0.59}  & \textbf{16.26} & $<$ \textbf{.001} & \textbf{.41} \\
    Velocity \( \times \) AVAS        & 1,23 &    & 0.40  &  .533 & .02 \\
    Velocity \( \times \) Robot & 2,46 & 0.96  &  0.11 & .892 & .00 \\
    AVAS \( \times \) Robot     & 2,46 & 1.00  &  1.07 & .351 & .04 \\
    Velocity \( \times \) AVAS \( \times \) Robot & 2,46 & 0.87  &  0.38 & .661 & .02 \\
    \hline
    \end{tabular}
    \caption{Results of rmANOVA on radial localization \textit{accuracy}. Significant effects highlighted in bold font.}
    \label{tab:results_radial}
    \end{table}

\subsection{Subjective assessment of the robot sounds}
    For the participants' subjective evaluation (annoyance, trust, valence) of the robot sounds with and without AVAS, we conducted a rmANOVA with the factors robot and AVAS, using a univariate approach and Huynh-Feldt correction for the degrees of freedom.

    The analyses consistently confirmed a significant and large effect of the robot type on the evaluations (annoyance: $F$(2,46)=89.55, $\tilde{\epsilon}$=0.80, $p<$.001, $\eta^2_p$=.80, trust: $F$(2,46)=16.39, $\tilde{\epsilon}$=0.78, $p<$.001, $\eta^2_p$=.42, valence: $F$(2,46)=50.37, $\tilde{\epsilon}$=0.82, $p<$.001, $\eta^2_p$=.69). We followed-up with two-sided pairwise $t$-tests with Bonferroni-correction. These showed that participants 1) perceived the HSR as significantly less annoying than the Turtlebot and the Go1, 2) trusted the HSR significantly more than the Turtlebot, and 3) evaluated the HSR significantly more positive than the Turtlebot and the Go1 (all $p_{\textit{Bonf}}<$.001, all $dz>$1.10). The ratings for the quadruped Go1 were descriptively better than for the Turtlebot, but the follow-up tests between these two robots confirmed only a significant difference regarding valence ($t$(23)=4.10, $p_{\textit{Bonf}}$=.004, $dz$=0.84).\\

    In sum, participants rated the HSR as less annoying, more trustworthy, and more positive regarding valence than the Turtlebot and Go1. They rated the Turtlebot and the Go1 as comparably annoying and trustworthy, but rated the Go1 as more positive than the Turtlebot.

    \begin{table}[t]
    \centering
    \begin{tabular}{rccccc}
    \hline
    \textit{\textbf{Predictor}} & \textit{\textbf{df}} & \(\mathit{\mathbf{\tilde{\epsilon}}}\) & \textit{\textbf{F}} & \textit{\textbf{p}} & \(\mathit{\mathbf{\eta^2_p}}\) \\
    \hline
    \textbf{Velocity}  & \textbf{1,23} &       &  \textbf{4.81} & \textbf{.039} & \textbf{.17} \\
    AVAS                       & 1,23 &       &  0.54 & .472 & .02 \\
    \textbf{Robot} & \textbf{2,46} & \textbf{0.76}  & \textbf{8.23} & \textbf{.003} & \textbf{.26} \\
    Velocity \( \times \) AVAS        & 1,23 &    & 0.11  &  .742 & .00 \\
    Velocity \( \times \) Robot & 2,46 & 0.93  &  0.58 & .550 & .02 \\
    AVAS \( \times \) Robot     & 2,46 & 0.93  &  1.28 & .288 & .05 \\
    Velocity \( \times \) AVAS \( \times \) Robot & 2,46 & 1.06  &  0.76 & .473 & .03 \\
    \hline
    \end{tabular}
    \caption{Results of rmANOVA on radial localization \textit{precision}. Significant effects highlighted in bold font.}
    \label{tab:results_SD_radial}
    \end{table}

\begin{figure}[t]
  \centering
 \includegraphics[width=0.99\linewidth]{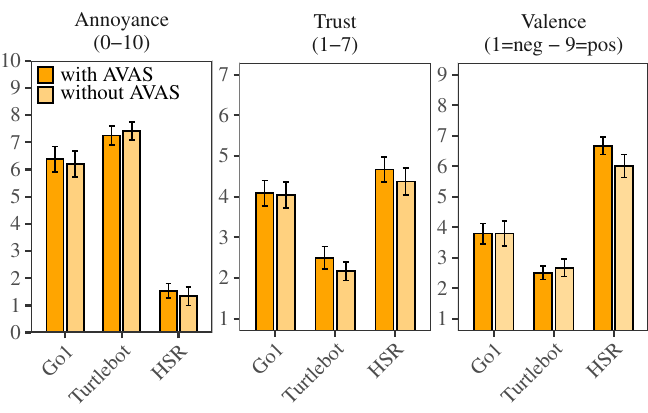}
 \caption{Mean subjective assessment of robots sounds with and without AVAS. Left: Annoyance. Middle: Trust. Right: Valence. Error bars indicate ±1 $SE$ of the 24 individual means.}
  \label{fig:resEval}
\end{figure}

\vspace{-0.5em}
\section{Conclusion}
\vspace{-0.25em}
\label{sec:conclusion}

    This study highlights the critical role of consequential sound in HRI, which demonstrates that the robot locomotion mechanism significantly influences the auditory localization of robots. While the quadruped Go1 and the small-wheeled Turtlebot robots were localized more accurately and precisely, the larger wheeled HSR--despite being rated as more trustworthy, less annoying, and more positive--was the most difficult for participants to localize, leading to the poorest performance. Overall, our findings underscore that a) consequential sound is a key factor for both subjective assessment and objective perception of mobile robots but that b) subjective and objective measures can be in clear contrast, i.e., the robot perceived as most pleasant (HSR) was also the hardest to localize. 
    This suggests that auditory information is essential for a favorable subjective evaluation \cite{allen_sound_2025}, while a good subjective evaluation does not necessarily ensure accurate auditory localization. 
    
     These findings allow for practical implications for the human-centered design of robots. The incongruence between participants' subjective experience and their objective perception establishes an interesting challenge for practical HRI design, in which both acceptance and safety in terms of robot localization are crucial for the successful integration of robots e.g., in public settings.

    The presented findings clearly support the need for robot manufacturers to consider robot's consequential sounds as a key factor in HRI, and to contemplate to supplement auditory robot design with an AVAS. However, the AVAS design should be optimized and evaluated in further studies.
    For social navigation algorithms, the results suggest that the robots' trajectory planning should account for the differences in the human auditory perception of different robot types to enhance the human predictability of robot motion in the sense of situational awareness. 
    From a psychological perspective, this study provides new insights into how humans perceive and evaluate robotic sounds, which highlight the importance of integrating both objective and subjective measures when designing effective and harmonious HRI.

\vspace{-0.5em}
\bibliographystyle{IEEEtran}
\bibliography{bibliography,bib_jorge,LocalizationPaper2025_updated}
\end{document}